\documentclass[a4paper]{article}
\usepackage{iwslt18,amssymb,amsmath,epsfig}
\def\reg{{\rm\ooalign{\hfil
     \raise.07ex\hbox{\scriptsize R}\hfil\crcr\mathhexbox20D}}}
\usepackage{color}

\usepackage{times}
\usepackage{latexsym}
\usepackage{arydshln}
\usepackage{url}
\usepackage{xcolor}
\usepackage{comment}
\usepackage{graphicx}
\usepackage{soul}
\usepackage{booktabs}
\usepackage{multirow}
\newcommand\sml[1]{\textcolor{black}{#1}}
\newcommand\short[1]{{}}
\newcommand\fix[1]{\textcolor{black}{#1}}

\title{Transfer Learning in Multilingual Neural Machine Translation \\with Dynamic Vocabulary}

 \makeatletter
 \def\name#1{\gdef\@name{#1\\}}
 \makeatother
\name{\em Surafel M. Lakew$^{\dagger +}$, Aliia Erofeeva$^{\dagger}$, Matteo Negri$^{+}$, Marcello Federico$^{ \star}$\thanks{(*) Work conducted while this author was at FBK.}, Marco Turchi$^{+}$}
\address{
$^{\dagger}$University of Trento, $^{+}$Fondazione Bruno Kessler, Trento, Italy \\ %$^{\star}$MMT Srl, Trento, Italy \\
$^{\star}$Amazon AI, East Palo Alto, CA 94303, USA \\
{\small \tt $^{\dagger}$name.surname@unitn.it, $^{+}$surname@fbk.eu, $^{\star}$marcfede@amazon.com}
}

\begin{document}
\maketitle
%
% * <surafelmelaku.lakew@unitn.it> 2018-08-31T20:50:32.814Z:
%
% ^.
% * <surafelmelaku.lakew@unitn.it> 2018-08-31T20:50:29.183Z:
%
% ^.
% * <surafelmelaku.lakew@unitn.it> 2018-08-31T20:50:17.167Z:
%
% ^.

%---------------
\begin{abstract}
We  propose a method to transfer knowledge across neural machine translation (NMT) models by means of a shared dynamic vocabulary.
Our approach allows to extend an initial model for a given language pair to cover new languages  by adapting its vocabulary  as long as new data become available (i.e., introducing new  vocabulary items if they are not included in the initial model).
The parameter transfer mechanism is evaluated in two scenarios:
\textit{i)} to adapt a trained single language NMT system to work with a new language pair and 
\textit{ii)} to continuously add new language pairs to grow to a multilingual NMT system.
In both the scenarios our goal is to improve the translation performance, while minimizing the training convergence time.
Preliminary experiments spanning five languages with different training data sizes (i.e., $5$k and $50$k parallel sentences) show a significant performance gain ranging from +$3.85$ up to +$13.63$ BLEU in different language directions. 
\sml{Moreover, when compared with training an NMT model from scratch, our transfer-learning approach allows us to reach higher performance after training up to $4$\% of the total training steps.}  
\end{abstract}

%---------------------
\section{Introduction}
Neural Machine Translation (NMT) has shown to surpass phrase based Machine Translation  
approaches not only in high-resource language settings, 
but also with low-resource~\cite{firat2016zero} and zero-resource translation tasks~\cite{johnson2016google,ha2016toward}. Although 
recent approaches yield promising results,  training models in low-resource settings remains a challenge for MT research \cite{koehn2017six}. 
\cite{johnson2016google} have shown that a multilingual NMT (M-NMT) model that utilizes a concatenation of data covering multiple language pairs (including high-resourced ones) 
can result in better performance in the low-resource translation task. Alternatively, \cite{zoph2016transfer} proposed a transfer-learning approach from an NMT ``\textit{parent-model}'' trained on a high-resource language to initialize a  ``\textit{child-model}'' in a low-resource setting showing consistent translation improvements on the latter task. 

Though effective, training models on a concatenation of data covering multiple language pairs or initializing them by transferring knowledge from a parent model does not consider the dynamic nature of new language vocabularies.
In relation to how and when model vocabularies are built, there can be two distinct scenarios. 
In the first one, all the training data for  all the language pairs are  available since the beginning. In this case, either  separate or joint sub-word segmentation models can be  applied on the training material to build  vocabularies that represent all the data~\cite{sennrich2015sub-word,wu2016google}. 
In the second scenario, training data covering different language directions are not  available at the same time (most real-world MT training scenarios fall in this category, in which new data or new needs in terms of domains or language coverage emerge over time).
In such cases, either: \textit{i)} new MT models are trained from scratch with new vocabularies built from the incoming training data, or \textit{ii)} the word segmentation rules  of a prior (parent) model are applied on the new data to continue the training as a fine-tuning task.
In all the scenarios, accurate word segmentation is crucial  to avoid  out-of-vocabulary (OOV) tokens. However, different strategies for the different training conditions can  result in longer training time or  performance degradations. 
More specifically, limiting the target task with the initial model vocabulary will result in: \textit{i)} a word segmentation that is unfavorable for the new language directions 
and \textit{ii)} a fixed vocabulary/\sml{model} dimension despite the varying language and training dataset size.

NMT models are not only data-demanding, but also require considerable time to be trained, optimized, and put into use. 
In particular real-word scenarios, strict time constraints prevent the possibility to deploy and use NMT technology (consider, for instance, emergency situations that require to promptly enable communication across languages~\cite{lewis2010haitian}). On top of this, when the available training corpora are limited in size, delivering  usable NMT systems (i.e., systems that can be used with the requirement of not making severe errors \cite{bentivogli:CSL2018})  
becomes prohibitive. 
In summary: \textit{i)} on the data side,  acquiring new training material for $x$ undefined languages is costly and not always possible, and \textit{ii)} on the model side, building an NMT system from scratch when new data become available raises efficiency and performance issues that are particularly relevant in low-resource scenarios.

We address these issues by introducing a method to  transfer knowledge across languages by means of a dynamic vocabulary. Starting from an initial model, our method allows to build new NMT models, either in a single or multiple language translation directions, by dynamically updating the initial vocabulary to new incoming  data.
For instance, given a trained German-English NMT system ($L_1$), the learned parameters can be transferred across models, while adopting new language vocabularies. In our experimental setting we test two transfer approaches: 
\begin{itemize}
\item {\tt progAdapt}: 
train a chain of consecutive M-NMT models by transferring the parameters of an initial model for $L_1$ to new language pairs $L_2$ \ldots $L_N$. In this scenario, the goal is to maximize performance on the new language pairs.   

\item {\tt progGrow}: 
progressively introduce  new language pairs to the initial model $L_1$  to create a growing M-NMT model covering $N$ translation directions. In this scenario, the goal is to maximize performance on all the language pairs.
\end{itemize}

Our experiments are carried out with  Italian$-$English, Romanian$-$English, and Dutch$-$English training data sets of different size, ranging from low-resource ($50$k) to extremely low-resource ($5$k) conditions.

As such, in a rather different way from previous work~\cite{zoph2016transfer}, we show our transfer-learning approach in a multilingual NMT model with \sml{dynamic vocabulary 
both in the source and target directions}. Our contributions are as follows: 
\begin{itemize}
\item we develop a \sml{transfer-learning} technique for NMT based on a dynamic vocabulary, 
which adapts the parameters learned on a parent task (language direction) to cover new target tasks;

\item through experiments in different scenarios, we show that our approach improves knowledge transfer across NMT models for different languages, particularly in low-resource conditions; 
\item we show that, with our transfer learning approach, it is possible to train a faster converging model that achieves better performance 
than a system trained from scratch. 
\end{itemize}

%UNCOMMENT IF SPACE IS AVAILABLE
\short{After summarizing  related work   on transfer-learning and multilingual MT in Section~2, in Section~3 we introduce  our  multilingual transfer-learning approach  and the two  settings in which it is evaluated (progressive adaptation to new language pairs and progressive growth to new translation directions). In Section~4, we describe the datasets, our baselines, and the models developed for our experiments. Then, our experimental results are presented and discussed in Section~5, before our conclusions in Section~6.}{}

%---------------------
\section{Related work}
\subsection{Transfer Learning}
Recent efforts~\cite{howard2018universal,radford2018improving} in natural language processing (NLP) research have shown promising results when transfer-learning techniques are applied to leverage existing models to cope with the scarcity of training data in specific domains or language settings. The advancements in NLP came  following a much larger impact of transfer-learning in computer vision tasks, such as classification and segmentation, either using features of ImageNet~\cite{sharif2014cnn} or by fine-tuning the last layers of a deep neural network~\cite{long2015fully}. 
Specific to NLP, pre-trained word embeddings \cite{qi2018andPre-trainedEmbd} used as input to the first layer of the network have become a common practice. In a broader sense, pre-trained models have been successfully exploited for several NLP tasks. \cite{mccann2017learned} used an MT model as a pre-training step to further contextualize word vectors for \sml{downstream} tasks like sentiment analysis, question classification, textual entailment, and question answering. In a similar way, a language model is utilized for pre-training in sequence labeling tasks~\cite{peters2017semi}, question answering, textual entailment, and sentiment analysis~\cite{peters2018deep}. 

Close to our approach, \cite{zoph2016transfer} explored techniques for transfer-learning across two NMT models. First, a ``parent'' model is trained with a large amount of available data. Then the encoder-decoder components are transferred to initialize the parameters of a low-resourced ``child'' model. In this parent-child setting, the decoder parameters of the child model are fixed at the time of fine-tuning. 
\short{This approach builds on the assumption that both the parent and the child models are translating into similar languages, whereas the encoder layers can be selectively updated.}{}Later, in \cite{nguyen2017transfer}, the parent-child approach has been extended to analyze the effect of using related languages on the source side.  

Although this work shares a related approach with~\cite{zoph2016transfer}, we diverge by our hypothesis not to selectively update only the encoder, allowing all the parameters to be updated as a beneficial strategy in our setting. Our strategy is based on both the source$\rightarrow$target and target$\rightarrow$source translation directions that we consider \sml{as transferable}. 
Moreover, our transfer-learning approach relies on a dynamic vocabulary that enforces changes in the trainable parameters of the network in contrast to fixing them\footnote{In future work, we plan to further study which parameters are more beneficial if transferred and which part of the network to selectively update.}.

%----------------------------
\subsection{Multilingual NMT}
In a one-to-many multilingual translation scenario, \cite{dong2015multi} proposed a multi-task learning approach that utilizes a single encoder for 
the source language and separate attention mechanisms and decoders for each  target language. \cite{luong2015multi} used distinct encoder and decoder networks for modeling 
multiple language pairs in a \textit{many-to-many} setting. Later, \cite{firat2016multi} introduced a way to share the attention mechanism across multiple languages. \sml{Aimed at avoiding translation ambiguities on the decoder side, a \textit{many-to-one} character level NMT setup \cite{lee2016fully} and a two/multi-source NMT \cite{zoph2016multi} were also proposed.} 
Inspired by \cite{sennrich2016controlling}, who automatically annotated the source side with artificial flags to manage the politeness level of the output, other works focused on controlling the grammatical voice \cite{yamagishi2016controlling}, the text domain \cite{Chen2016,Chu2017}, and enforcing gender agreement \cite{Elaraby2018}.
Simplified yet efficient multilingual NMT approaches have been proposed by \cite{johnson2016google} and \cite{ha2016toward}. The approach in \cite{ha2016toward} applies a language-specific code to words from different languages in a mixed-language vocabulary. 
The approach in \cite{johnson2016google}, by prepending a {\it language flag} to the input string, greatly simplified multilingual NMT eliminating the need of having separate encoder/decoder networks and attention mechanism for each new language pair. 
In this work we follow a similar strategy by incorporating an artificial language flag.

%-----------------------------------
\section{Transfer Learning in M-NMT}
\label{sec:TL-in-M-NMT}
In this work, we cast transfer-learning in a multilingual neural machine translation (M-NMT) task as the problem of dynamically changing/updating the vocabulary of a trained NMT system.
 In particular, transfer-learning across models is assumed to: 
\textit{i)} include a strategy to add new language-specific items to  an existing NMT vocabulary, and \textit{ii)} be able to manage a number of new translation directions in different transfer rounds, either by covering them one at a time (i.e., in a chain where new languages are covered stepwise)  or simultaneously (i.e., pursuing all directions at each step).    
 Our investigation focuses on two aspects. The first one is how the parameters of an existing model can be transferred to a target one for a new language pair.
The second aspect is how to limit the impact of parameters' transfer on the performance of the initial  model as long as new language directions are added. \sml{For convenience, we refer to our approach as {\tt TL-DV} ({\it Transfer-Learning using Dynamic Vocabulary})}.

\begin{figure*}[!t]
\centerline{\epsfig{figure=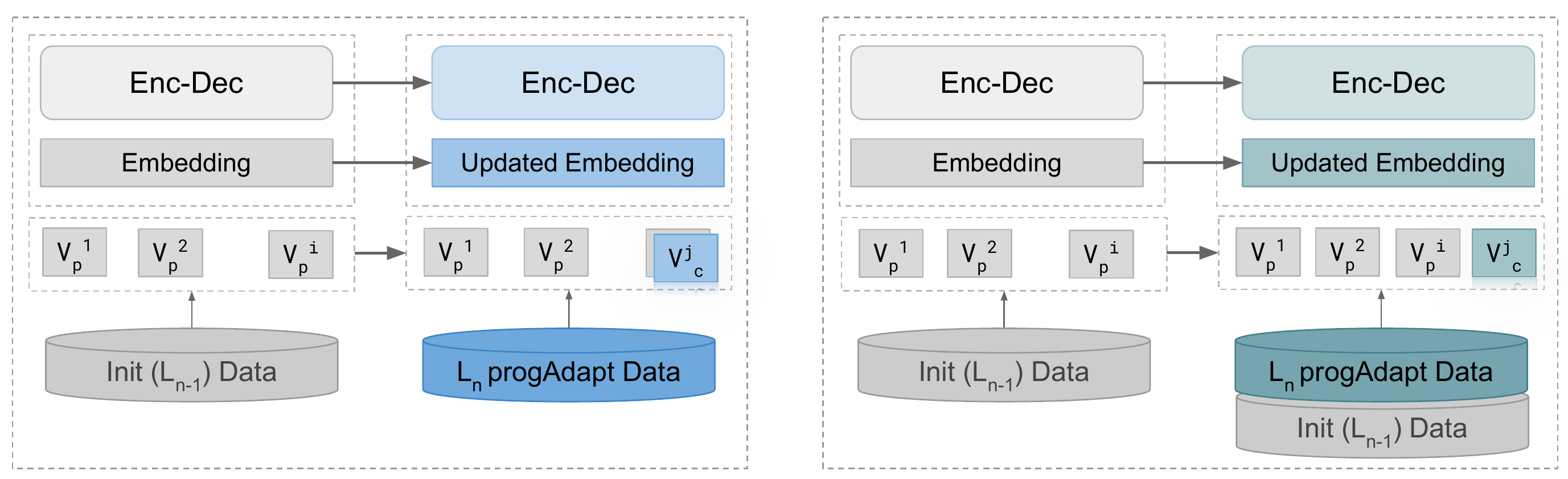,width=5.2in}}
\caption{{\it Transfer-Learning; (left) from an initial NMT model to a new language pair, model is applied after inserting the new vocabulary entries, for instance, the initial model \textcolor{gray}{$L_{n-1}$} parameters are transfered to \textcolor{black}{$L_n$} with the updated embedding space (i.e., keeping $V_p^1$, $V_p^2$ as overlapping entries, while replacing the non-overlapping $V_p^i$ with $V_c^j$ new language vocabularies), 
and (right) from an initial model \textcolor{gray}{$L_{n-1}$} to \textcolor{black}{$L_{n}$}, but incorporating both the previous and new language pair data and vocabulary entries.}}
\label{tl-dv}
\end{figure*}

As shown in Figure~\ref{tl-dv}, our transfer-learning approach is evaluated in two conditions: 
\begin{itemize}
\item {\tt progAdapt}, in which progressive updates are made on the assumption that new target NMT task data become available for one language direction  at a time (i.e., new language directions are covered sequentially). In this condition, our goal is to maximize performance on the new target tasks by taking advantage of parameters learned in their parent task;
\item {\tt progGrow}, \sml{in which progressive updates are made on the same assumption of receiving new target task data as in {\tt progAdapt}, but with the additional goal of preserving the performance of the previous language directions.} 
\end{itemize}
We discuss these two scenarios below in $\S$\ref{progressiveAdapt} and $\S$\ref{progressiveGrowth}.

%------------------------------
\subsection{Dynamic Vocabulary}\label{sect:TL-DV}
In the defined scenarios, we update the vocabulary $V_p$ of the previous model with the current language direction vocabulary $V_c$. 
The approach simply keeps the intersection (same entries) between $V_p$ and $V_c$, 
whereas replacing $V_p$ entries with $V_c$ if the entries of the former vocabulary do not exist in the latter. At training time, these new entries are randomly initialized, while the 
intersecting items maintain \fix{the embeddings of the former model}.
\sml{The alternative approach to dynamic vocabulary in a continuous model training is to use the initial model vocabulary $V_p$, which we refer to as static-vocabulary.}

%--------------------------------
\short{\subsection{Initial Model}
\label{init_model}
Our initial model is an NMT system trained with the concatenation of source and target data to create a bi-NMT model. 
For instance, for the German-English corpus, the initial model is trained by aggregating German$\rightarrow$English and English$\rightarrow$German data in one model.
The initial  model is assumed to be trained with a high-resourced language pair. In fact, the {\tt init} model trained as M-NMT (i.e., two or more directions) allows to capture more than one language features and vocabularies on both the encoder and decoder sides, consequently increasing the transferable hidden representations and vocabularies to the subsequent transfer tasks.
\fix{In this work, we follow a similar strategy as in~\cite{johnson2016google} by incorporating an artificial language flag.}
}{}

%--------------------------------------------------
\subsection{Progressive Adaptation to New Languages}
\label{progressiveAdapt}
In this scenario, starting from the {\tt init} model ($L_1$), we perform  
progressive adaptation by initializing the training of a model at each step ($L_n$) with the previous model  ($L_{n-1}$). At time of reloading the model from $L_{n-1}$,
 a {\tt TL-DV} update is performed as described in $\S$\ref{sec:TL-in-M-NMT}. In this approach, the dataset of the initial model is not included at the current training stage. This allows the adaptation to the new language without unnecessary word segmentation that may arise by applying the initial model's segmentation rules. 
As shown in Figure~\ref{tl-dv} (left), the adaptation on any of the $L_n$ stages is language independent, though subject to the available training dataset. 
We refer to the application of this approach in the experimental settings and discussion as {\tt progAdapt}.

%--------------------------------------------------------
\subsection{Progressive Growth of Translation Directions}
\label{progressiveGrowth}
In this scenario, an initial model $L_1$ is simultaneously adapted to an incremental number of translation directions, under the constraint that the level of performance on $L_1$ has to be maintained.
For a simplified experimental setup, we will incorporate a single language pair (source$\rightarrow$target) at a time, when adapting to $L_n$ from $L_{n-1}$ (see Figure~\ref{tl-dv} (right)).
We refer to the application of this approach in the experimental settings and discussion as {\tt progGrow}.

%------------------------------------------------
\section{Experimental Setting}\label{experiments}
\subsection{Dataset and Preprocessing}
Our experimental setting includes the {\tt init} model language pair (German-English) and three additional language pairs (Italian-English, Romanian-English, and Dutch-English) for testing the proposed approaches. We use publicly available datasets from the WIT$^3$ TED corpus~\cite{cettoloEtAl:EAMT2012}. Table~\ref{table:dataset} shows the summary of the training, dev, and test sets. To simulate an extremely low-resource ($M_{ELR}$) and low-resource ($M_{LR}$) model settings,  $5$K and $50$K sentences are sampled from the last three language pairs' training data.

At the preprocessing step, we first tokenize the raw data and remove sentences longer than 70 tokens. As in~\cite{johnson2016google}, we prepend a ``language flag'' on the source side of the corpus for all multilingual models. For instance, if a German source is paired with an English target, we append {\tt <2ENG>} at the beginning of source segments\short{\footnote{The ``language flag'' is chosen following the {\tt ISO-639} three letter code for language names \url{ https://en.wikipedia.org/wiki/List_of_ISO_639-2_codes}}}{}. 
Next, a shared byte pair encoding (BPE) model~\cite{sennrich2015sub-word} is trained using the union of the source and target sides of each language pair. 
Following \cite{denkowski2017stronger}, the number of BPE segmentation rules is set to $8,500$ for the data size used in our experimental setting. 
At different levels of training ($L_i$), a BPE model with respect to the language pairs is then used to segment the training, dev, and test data into sub-word units.
\fix{While, the vocabulary size of the {\tt init} is fixed, the vocabulary varies in the consecutive training stages depending on the overlap of sub-word units and lexical similarity between two language pairs.}

\begin{table}
\caption{\label{table:dataset} {\it Languages and dataset sizes for train, dev, and test sets of the {\tt init} model for De-En direction and other pairs assumed to be received progressively (It-En, Ro-En, Nl-En).}}
\vspace{2mm}
\centerline{
\begin{tabular}{lrrrr}
\hline
Language 				& Train 	& Dev 	& Test 	& Received \\
\hline  %\hline
German(De)-En 	& 200k  	& 1497 	& 1138 	& {\tt init} \\ 
\hdashline
Italian(It)-En 	&  5k/50k 	& 1501 	& 1147 	& {\tt $L_2$ } \\
Romanian(Ro)-En 	&  5k/50k 	& 1633 	& 1129 	& {\tt $L_3$ } \\
Dutch(Nl)-En 	&  5k/50k 	& 1726 	& 1181 	& {\tt $L_4$ } \\
\hline
\end{tabular}}
\end{table}

%---------------------------------
\subsection{Experimental Settings}
All systems are trained using the Transformer~\cite{vaswani2017attention} model implementation of the OpenNMT-tf sequence modeling framework\footnote{https://github.com/OpenNMT/OpenNMT-tf}~\cite{klein2017opennmt}. At training time, to alternate between dynamic and static vocabulary, we utilized an updated version of the script within the same framework. 
For all trainings, we use LazyAdam, a variant of the Adam optimizer~\cite{kingma2014adam}, with an initial learning rate constant of $2$ and a dropout~\cite{srivastava2014dropout,gal2016theoreticallyDropout} of $0.3$. The learning rate is increased linearly in the early stages (\emph{warmup\_training\_steps=$16,000$}), and afterwards it is decreased with an inverse square root of the training step.
 
To train our models using Transformer, we employ a uniform setting with $512$ hidden units and embedding dimension, and 6 layers of self-attention encoder-decoder network. The training batch size is of $4096$ sub-word tokens. At inference time, we use a beam size of $4$ and a batch size of $32$. Following~\cite{vaswani2017attention} and for a fair comparison, all baseline experiments are run for 100k training steps, i.e., all models are observed to converge within these steps. The consecutive experiments converge in variable training steps. However, to make sure a convergence point is reached, all restarted experiments on $L_i$ are run for additional $50$K steps. All models are trained on a {\tt GeForce-GTX-1080} machine with a single GPU. 
\fix{Systems are compared in terms of  BLEU~\cite{papineni2002bleu} using the \emph{multi-bleu.perl} implementation\footnote{A script from the Moses SMT toolkithttp://www.statmt.org/moses}, on the single references of the official IWSLT test sets.}

%---------------------------
\subsection{Baseline Models}\label{subsec:baselines}
To evaluate and compare with our approach, we train single language pair baseline models corresponding to the newly introduced language pairs at each $L_i$ training stage.  
The baseline models, referred to as {\tt Bi-NMT}, are  separately trained from scratch in a bi-directional setting (i.e., source $\leftrightarrow$ target). 
\fix{In addition, we report scores from a multilinugal ({\tt M-NMT}) model trained with the concatenation of all available data in each training stage. 
} 
The alternative baseline are built by fine-tuning the {\tt init} model. These models use the vocabulary (word segmentation rules) of the {\tt init} model, avoiding the proposed dynamic vocabulary. This fine-tuning approach is prevalent in continued model trainings, \fix{for adapting NMT models \cite{michel2018extreme,vilar2018learning} or improving zero-shot and low-resource translation tasks \cite{lakew2017improving,lample2017unsupervised,gu2018universal}.}
\sml{For the alternative baseline where we fine-tune {\tt init} with its static-vocabulary, we observed that results were mostly analogous to {\tt Bi-NMT} models. Hence, we avoided this comparison in this work and relied on the former baselines.
}

%-------------------------------
\section{Results and Discussion}
Experiments are performed using the {\tt progAdapt} (see $\S$\ref{progressiveAdapt}) and {\tt progGrow} ($\S$\ref{progressiveGrowth}) approaches. The experimental results with the associated discussion are presented in Table~\ref{table:low-resource} for models characterized by relatively low-resource data ($M_{LR}$), and in Table \ref{table:ext-low-resource} for an extremely low-resource condition ($M_{ELR}$). In both dataset conditions, the performance of the proposed approaches is compared with the baseline systems ({\tt Bi-NMT} and \fix{{\tt M-NMT}}, see $\S$\ref{subsec:baselines}). 

The {\tt init} model which is trained with a data size $4$X larger than $M_{LR}$ and $40$X the size of $M_{ELR}$, achieves BLEU scores of $26.74$ and $23.30$, respectively, for the De-En and En-De directions. In Table \ref{table:low-resource} and \ref{table:ext-low-resource}, the {\tt progAdapt} is reported for each training stage (i.e., $L_2$, $L_3$, and $L_4$), whereas the {\tt progGrow} is reported for the final stage $L_4$. 
\fix{ Moreover, Table \ref{table:related-lang} analyzes the effect of language relatedness and training stage reordering in our {\tt TL-DV} approach. Bold highlighted BLEU scores show the best performing approach, while the $^{\uparrow\downarrow}$ arrows indicate statistically significant differences of the hypothesis against the better performing baseline ({\tt M-NMT}) using bootstrap resampling ($p<0.05$)~\cite{Koehn2004}.}

%--------------------------------
\subsection{Low-Resource Setting}
\begin{table}[!t]
\caption{\label{table:low-resource} {\it $M_{LR}$ models performance i)~at $L_1$ for the {\tt init} De-En direction and baseline ({\tt Bi-NMT}) It-En, Ro-En, and Nl-En directions, ii)~at $L_{2/3/4}$ for {\tt progAdapt}, and iii)~at $L_4$ for the {\tt progGrow} approach.}} 
\vspace{2mm}
\begin{tabular}{@{}lccccc@{}}
\toprule
													& Dir						& De-En                 & It-En                    & Ro-En                   & Nl-En                    \\ \midrule
\multirow{2}{*}{{\tt Init/Bi-NMT}} 		 & \textgreater{}		  & {\bf 26.74 }	      & 25.21                   & 10.80                    & 21.75                    \\  
                               						& \textless{}            & 23.30                  & 22.39                   & 12.94                    & 19.75                    \\ \cmidrule(l){2-6} 
\multirow{2}{*}{{\tt M-NMT}}  			 & \textgreater{}        & 24.14 	         	  & 26.42                   & 22.17                    & 24.00                    \\  
                               						& \textless{}          	& 21.80                  & 23.57                   & 17.35                    & 21.25                    \\ \cmidrule(){1-6} 
\multirow{2}{*}{{\tt ProgAdapt}}    	& \textgreater{}      	& -                        	& $^{\uparrow}${\bf 30.08}            & $^{\uparrow}${\bf 24.43}              & $^{\uparrow}${\bf 26.36} \\ 
                               					   & \textless{}          	& -                 		& $^{\uparrow}${\bf 26.24}            & $^{\uparrow}${\bf 20.31}              & $^{\uparrow}${\bf 25.52} \\ \cmidrule(l){2-6} 
{\tt ProgGrow}                      		 & \textgreater{}      	 & 26.22                  & $^{\uparrow}${\bf 29.61}            & {\bf 23.23}              & {\bf 24.78}   \\ \bottomrule
\end{tabular}
\end{table}

For each language pair (i.e., It-En, Ro-En, and Nl-En), the results of the baseline models {\tt Bi-NMT}  trained using the available $50$K parallel data ($M_{LR}$ setting) are  
presented in the first two rows of Table \ref{table:low-resource}. The {\tt progAdapt} results are reported from three consecutive adaptations to new language directions. These include the {\tt init} to It-En, followed by the adaptation to Ro-En, and then to Nl-En. Compared to the corresponding {\tt Bi-NMT} and \fix{M-NMT} models, all of the three progressive adaptations using the dynamic vocabulary technique achieved a higher performance gain.

If we look at the specific level of adaption ($L_i$) \fix{against the {\tt Bi-NMT}}, we observe that the It-En direction showed a +4.87 and +3.85 gain for the En and It target, respectively. When we take this model and continue the adaptation to Ro-En and Nl-En, we see a similar trend where the highest gain is observed on $L_3$ for the Ro-En direction with +13.63 and +7.37 points. 
These significant improvements over the baseline models tell us that transfer-learning using dynamic vocabulary in a multilingual setting is a viable direction. Its capability to quickly tune the representation space of the {\tt init} model to deliver improved results is an indication of the importance of using different word representations for each language pair\footnote{We reserve the adaptation from the {\tt init} model directly to all the three new language pairs and the comparison with the current setting for future work.}.

In case of the {\tt progGrow}, \short{where we focused our experiments to the English (En) directions only,}{} we observed a similar improvement trend as in the {\tt progAdapt} approach. The results are reported from the final stage ($L_4$) of the model growth, but improvements are consistent throughout the {\tt $L_2$} and {\tt $L_3$} stages. 
\fix{The {\tt M-NMT} outperformed the {\tt Bi-NMT} models except for De-En pair. However, compared to the multilingual model as an alternative method for achieving cross-lingual transfer-learning, our approach shows improvements in the consecutive training stages.}
Overall, our observation is that the suggested {\tt progGrow} model can accommodate new translation directions when the data are received. Most importantly, improvements are observed for these newly introduced languages without altering the performance of the {\tt init} model in the De-En direction. 

Specific to each language direction, It-En shows a comparable performance with the {\tt progAdapt} approach, whereas in case of Ro-En and Nl-En a small degradation ranging from $0.47$ (De-En) to $1.58$ (Nl-En)
is observed. The loss in performance is likely due to the increased ambiguities in the encoder side of the {\tt progGrow} model, where at both training and inference time there does not exist a disambiguation mechanism between languages except the prepended language flag. This observation, which sheds a light on our initial expectation of more data aggregation benefiting the model performance,  requires further investigation.

%------------------------------------------
\subsection{Extremely Low-Resource Setting}
\begin{table}[!t]
\caption{\label{table:ext-low-resource} {\it $M_{ELR}$ models performance i) at $L_1$ for the {\tt init} De-En direction and baseline ({\tt Bi-NMT}) It-En, Ro-En, and Nl-En directions, ii) at $L_{2/3/4}$ for {\tt progAdapt}, and iii) at $L_4$ for the {\tt progGrow} approach.}}
\vspace{2mm}
\begin{tabular}{@{}lccccc@{}}
\toprule
													& Dir           					& De-En                    & It-En                       & Ro-En                    & Nl-En                    \\ \midrule
\multirow{2}{*}{{\tt Init/Bi-NMT}}		 & \textgreater{}					& {\bf 26.74}				& 7.64                     & 4.56                     & 5.69                     \\  
                               						& \textless{}      					& 23.30                    	& 5.25                     & 3.86                     & 5.14                     \\ \cmidrule(l){2-6} 
\multirow{2}{*}{{\tt M-NMT}}  			& \textgreater{}       				& 24.96		         	& {\bf 16.26}         & {\bf 12.67}             & {\bf 15.59}                    \\ 
                               						& \textless{}			        	& 21.67               	& 10.38                  	& 8.67                    & 12.72                    \\ \cmidrule(){1-6}
\multirow{2}{*}{{\tt ProgAdapt}}      & \textgreater{}  				& -                        & $^{\downarrow}$15.16	              & $^{\downarrow}$11.03                & $^{\downarrow}$11.52                     \\ 
                               					 & \textless{}			      		& -                        & $^{\uparrow}${\bf 14.40 }    	  & $^{\uparrow}${\bf 11.10 }       & {\bf 13.57 }             \\ \cmidrule(l){2-6} 
{\tt ProgGrow}                       	   & \textgreater{}  				& 25.61                  & $^{\downarrow}$15.02                &  $^{\downarrow}$11.20        	    & $^{\downarrow}$13.56                   \\ \bottomrule
\end{tabular}
\end{table}

In a similar way with what we observed in the $M_{LR}$ experiments, the baseline models in the extremely low-resource setting demonstrate poor performance. Looking at our approaches, we observe a relatively higher gain at the first stage of {\tt progAdapt} and {\tt progGrow}. For instance, for the It-En pair there is a +7.52 improvement compared to the +4.87 in the $M_{LR}$ models (see Table~\ref{table:low-resource}) \fix{over the {\tt Bi-NMT} model}. In the subsequent additional language directions (i.e., Ro-En and Nl-En), we also observe a similar trend. \fix{However, in comparison with the {M-NMT}, both of our approach perform poorly when translating to the En target. The main reason for this could be the aggregation of all the available data for a single run in the {\tt M-NMT} model, while our approaches exploit data when it becomes available in a continuous training. Alternatively the distance between each language pair could play a significant role when we adapt in an extremely sparse data.}

\noindent
{\bf prog-Adapt/Grow with Related Languages}. \fix{When related language pairs are consecutively added ({\tt $L_{n-1}$} and {\tt $L_n$}) at each training stages, our {\tt TL-DV} approach showed the best performance. For instance, for the Nl-En experiments, we changed the sequence of the added language pair moving from a random order to a sequence based on the similarity to the {\tt init} model.}
\begin{table}[!t]
\caption{\label{table:related-lang} {\it $M_{LR}$ and $M_{ELR}$ models performance at $L_1$ for {\tt progAdapt} and {\tt progGrow} approaches in a closely related De-En ({\tt init}) and Nl-En language pairs setting.}}
\vspace{2mm}
\begin{tabular}{@{}lccccc@{}}

																					  & & \multicolumn{2}{c}{{\tt $M_{LR}$}} & \multicolumn{2}{c}{{\tt $M_{ELR}$}} \\ \cmidrule(l){2-6}
													& Dir					 		& De-En                 & Nl-En                  					& De-En                   		& Nl-En                    \\ \midrule
\multirow{2}{*}{{\tt ProgAdapt}}    	& \textgreater{}      		& -                        	& $^{\uparrow}${\bf 27.23}             & 					              & {\bf 16.21} \\ 
                               					   & \textless{}          		& -                 		& $^{\uparrow}$25.51		             &               				  & {\bf 15.86} \\ \cmidrule(l){2-6} 
{\tt ProgGrow}                      		 & \textgreater{}      	 	& 26.62                  & $^{\uparrow}${\bf 26.41}               & 26.52		              	& $^{\uparrow}${\bf 15.52}   \\ \bottomrule
\end{tabular}
\end{table}
\fix{Table~\ref{table:related-lang} shows the results from {\tt progAdapt} and {\tt progGrow}, when the Nl-En pair is used at the $L_1$ training stage. The {\tt $M_{LR}$} results confirm the trend observed in Table \ref{table:low-resource}, however, with a relatively better performance when translating in to English. Most importantly, the {$M_{ELR}$} results show a consistent and larger gain of $+4.69$ (Nl-En) and $+2.29$ (En-Nl) with the {\tt progAdapt}, and $+1.96$ (Nl-En) with {\tt progGrow} compared to the corresponding results in Table \ref{table:ext-low-resource}. Thus, we emphasize on the degree of language similarity as a direct influencing factor when incorporating a new language pair both in {\tt progAdapt} and {\tt progGrow} approaches.
}

\begin{figure}[!t]
\centerline{\epsfig{figure=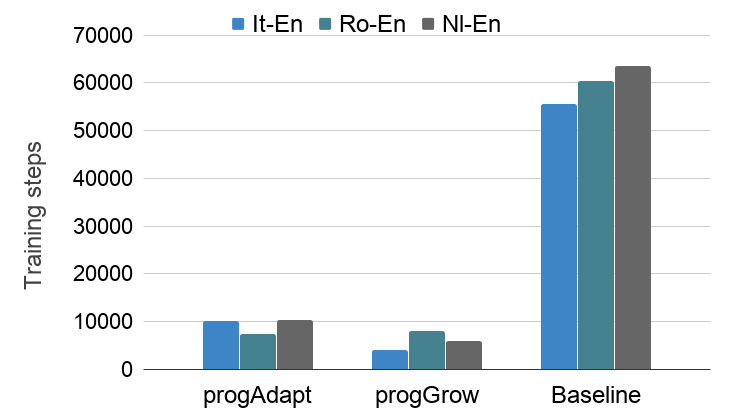,width=3.0in}}
\caption{{\it Model training steps number comparison for the three different language pairs between the baseline (rightmost) and the proposed approaches \fix{in the $M_{ELR}$ setting}.
}}
\label{tl-dv-convergence-5k}
\end{figure}

\noindent
{\bf Prog-Adapt/Grow with Faster Convergence}. 
The other main advantage of our {\tt TL-DV} approach comes from the time  a model takes to restart from the {\tt init} model and reach a convergence point with better performance.
\short{In all experiments with our dynamic vocabulary, a converged model is found within $4$ to $20$K additional training steps.}{} 
\fix{In all experiments with our {\tt TL-DV} approach a converged model is found within $10$K steps for $M_{ELR}$ and $20$K for $M_{LR}$ training settings.} Compared to $\approx$100K steps needed by a model trained from scratch to reach good performance, our approach takes only 4\% to 20\% of training steps with significantly higher performance. For instance, taking into consideration the $M_{ELR}$ models, Figure \ref{tl-dv-convergence-5k} illustrates the steps required for the baseline systems to converge (Table~\ref{table:ext-low-resource}), in comparison with our approach where {\tt progGrow} shows to converge slightly faster than {\tt progAdapt}. However, with the relatively larger data of the $M_{LR}$ models, the {\tt progAdapt} approach proves to converge much faster than {\tt progGrow}, for the reason that the newly introduced vocabulary and training dataset sizes are smaller compared to the concatenation of the {\tt init} and $L_{i}$ data.

\begin{figure}[!t]
\centerline{\epsfig{figure=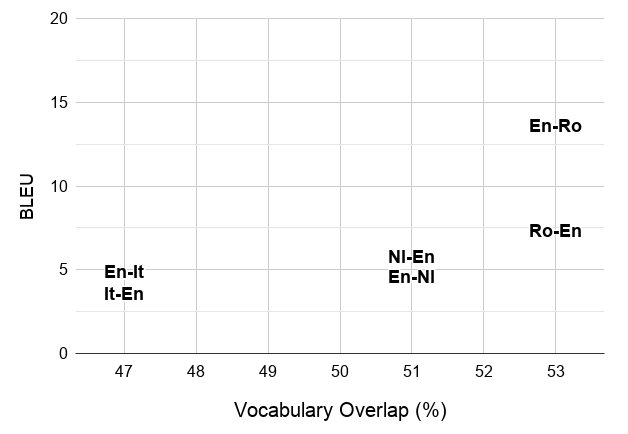,width=2.9in}}
\caption{{\it The difference in performance between the baseline and {\tt progAdapt} models (Tgt$\rightarrow$Src and Src$\rightarrow$Tgt directions)
in relation with the shared vocabulary between model $L_i$ and new language pair model $L_{i+1}$.} }
\label{figure:tl-dv-vocab}
\end{figure}

\sml{We further analyzed the influence of shared vocabularies between models $L_i$ and $L_{i+1}$ on the performance of {\tt TL-DV}. For this discussion, we took the {\tt progAdapt} $M_{LR}$ model from all stages. Figure~\ref{figure:tl-dv-vocab} summarizes the improvement differences from consecutive models in relation to the percentage of shared vocabularies. 
For instance, {\tt init} and the $L_2$ (It-En) model vocabularies have a $47$\% overlap, whereas $L_3$ and $L_4$ share $53$\% and $51$\% with the previous model. The interesting aspect of the shared vocabulary comes from the  increase in model performance with a higher fraction of shared vocabulary entires. Thus, a larger number of shared parameters between two consecutive models allows for a better gain in performance of the latter. 
}

\sml{The results achieved by the transfer-learning with dynamic vocabulary approach in two different training size conditions show that: {\it i)} adapting a trained NMT model to a new language pair improves  performance on the target task significantly, and {\it ii)}  it is possible to train a model faster to achieve better performance. Overall, the capability of injecting new vocabularies for new language pairs  in the initial model is a crucial aspect for efficient and fast adaptation steps.}

%--------------------
\section{Conclusions}
In this work, we proposed a transfer-learning approach within a multilingual NMT. Experimental results show that our dynamic vocabulary based transfer-learning improves model performance in a significant way of up to $9.15$ in an extremely low-resource and up to $13.0$ BLEU in a low-resource setting over a bilingual baseline model.

In future work, we will focus on finding the optimal way of transferring\short{the {\tt init}'s trainable}{ model} parameters\short{into the consecutive models}. Moreover, we plan to test our approach for various languages and language varieties.

%--------------------------
\section{Acknowledgments}
This work has been partially supported by the EC-funded project ModernMT (H2020 grant agreement no. 645487). We also gratefully acknowledge the support of NVIDIA Corporation with the donation of the Titan Xp GPU used for this research. Moreover, we thank the Erasmus Mundus European Program in Language and Communication Technology.
%--------------------------

\bibliography{bib.bib}
\bibliographystyle{IEEEtran}
\end{document}